\documentclass{article}
\usepackage{spconf,amsmath,graphicx}
\usepackage{amsmath,bm}
\usepackage{hyperref}
\usepackage{multirow}
\usepackage{amsfonts}
\usepackage{booktabs,tabulary}
\usepackage{threeparttable}
\usepackage{float}
\usepackage{subcaption}
\usepackage{verbatim}
\usepackage[noend]{algpseudocode}
\usepackage{algorithm}
\usepackage{adjustbox}
\usepackage{graphicx}
\usepackage{lipsum}
\usepackage{diagbox} 
\usepackage{wrapfig}
\usepackage{amssymb}

\newcommand{\HH}{{\mathcal H}}
\newcommand{\RR}{{\mathbb R}}
\newcommand{\bx}{{\pmb x}}

\newcommand{\balpha}{{\pmb \alpha}}

\newcommand\defeq{\triangleq}
\newcommand\norm[1]{\left\lVert#1\right\rVert}
\newcommand\tn[1]{\textnormal{#1}}

\algnewcommand\algorithmicswitch{\textbf{switch}}
\algnewcommand\algorithmiccase{\textbf{case}}
\algnewcommand\algorithmicassert{\texttt{assert}}
\algnewcommand\Assert[1]{\State \algorithmicassert(#1)}%
\algdef{SE}[SWITCH]{Switch}{EndSwitch}[1]{\algorithmicswitch\ #1\ \algorithmicdo}{\algorithmicend\ \algorithmicswitch}%
\algdef{SE}[CASE]{Case}{EndCase}[1]{\algorithmiccase\ #1}{\algorithmicend\ \algorithmiccase}%
\algtext*{EndSwitch}%
\algtext*{EndCase}%

\expandafter\def\expandafter\normalsize\expandafter{%
    \normalsize
    \setlength\abovedisplayskip{8pt}
    \setlength\belowdisplayskip{8pt}
    \setlength\abovedisplayshortskip{8pt}
    \setlength\belowdisplayshortskip{8pt}
}

\title{KERNEL MACHINES BEAT DEEP NEURAL NETWORKS ON MASK-BASED SINGLE-CHANNEL SPEECH ENHANCEMENT }
\name{Like Hui, Siyuan Ma, Mikhail Belkin}

\address{Department of Computer Science and Engineering, The Ohio State University, USA\\\small{hui.87@osu.edu, \{masi, mbelkin\}@cse.ohio-state.edu}}
\begin{document}
%

\maketitle
\begin{abstract}
We apply a fast kernel method for mask-based single-channel speech enhancement.
Specifically, our method solves a kernel regression problem associated to a non-smooth kernel function (exponential power kernel) with a highly efficient iterative method (EigenPro).
Due to the simplicity of this method, its hyper-parameters such as kernel bandwidth can be automatically and efficiently selected using line search with subsamples of training data.
We observe an empirical correlation between the regression loss (mean square error) and regular metrics for speech enhancement.
This observation justifies our training target and motivates us to achieve lower regression loss by training separate kernel model per frequency subband.
We compare our method with the state-of-the-art deep neural networks on mask-based HINT and TIMIT.
Experimental results show that
our kernel method 
consistently outperforms deep neural networks while requiring less training time.

\end{abstract}
\begin{keywords}
large-scale kernel machines, deep neural networks, speech enhancement, exponential power kernel, automatic hyper-parameter selection
\end{keywords}

\section{Introduction}
\label{sec:intro}

The challenging problem of single-channel speech enhancement has received significant attention  in research and applications. 
In recent years the dominant methodology for addressing single-channel speech enhancement has been based on neural networks of different architectures~\cite{wang2018supervised,zhang2018deep}. Deep Neural Networks (DNNs) present an attractive learning paradigm due to their empirical success on a range of problems and efficient optimization. 

In this paper, we  demonstrate that modern large-scale kernel machines are a powerful alternative to DNNs, capable of matching and surpassing their performance while utilizing less computational resources in training.  
Specifically, we take the approach to speech enhancement based on the Ideal Binary Mask (IBM) and Ideal Ratio Mask (IRM) methodology. The first application of DNNs to this problem was presented in \cite{wang2013towards}, which used a DNN-SVM (support vector machine) system to solve the classification problem corresponding to estimating the IBM.~\cite{wang2014training} compared different training targets including IRM.~\cite{xu2015regression} proposed a regression-based approach to estimate speech log power spectrum. Recently, 
\cite{wang2017recurrent} applies recurrent neural networks to similar mask-based tasks and \cite{pandey2018new} applies convolutional networks to the spectrum-based tasks.



Kernel-based shallow models (which can be interpreted as  two-layer neural networks with a fixed first layer), were also proposed to deal with speech tasks. In particular,~\cite{huang2014kernel} gave a kernel ridge regression method, which matched DNN on TIMIT. Inspired by this work,~\cite{chen2016efficient} applied an efficient one-vs-one kernel ridge regression for speech recognition. 
\cite{lu2016comparison} developed kernel acoustic models for speech recognition.
Notably, these approaches require large computational resources to achieve performance comparable to neural networks.

In our opinion, the computational cost of scaling to larger data has been a major factor limiting the success of these methods. In this work we apply a recently developed highly efficient  kernel optimization method EigenPro~\cite{ma2018learning}, which allows kernel machines to handle  large datasets.

We conduct experiments on standard datasets using mask-based training target. Our results show that, with EigenPro iteration, kernel methods can consistently outperform the performance of DNN 
in terms of the target mean square error (MSE) as well as the commonly used speech quality evaluation metrics including perceptual evaluation of speech quality (PESQ) and short-time objective intelligibility  (STOI).

\begin{figure*}[!htb]
~~~~~~~~~~~~~
\includegraphics[width=\textwidth]{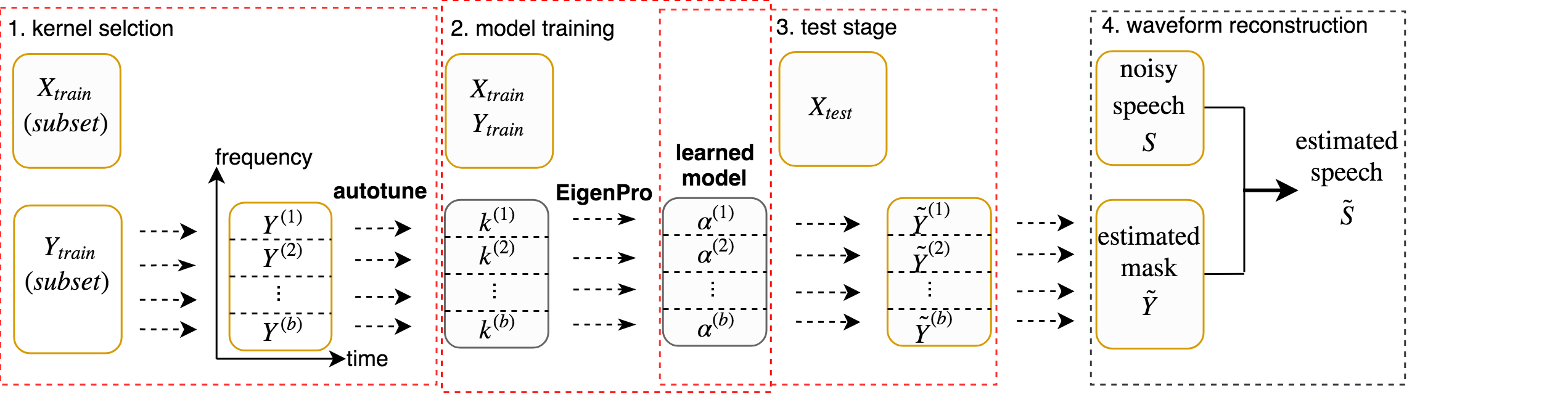}
\vspace{-8mm}
\caption{Kernel-based speech enhancement framework}
\label{fig:framework}
\end{figure*} 

The contributions of our paper are as follows:
\begin{enumerate}
\vspace{-3mm}
\itemsep-0.4em 
\item Using modern kernel algorithms we show performance on mask-based speech enhancement  surpassing that of neural networks and requiring less training time. 
\item To achieve the best performance, we use exponential power kernel, which, to the best of our knowledge, has not been used for regression or classification tasks.
\item  The simplicity of our approach
allows us to develop a nearly automatic hyper-parameter selection procedure based on target speech frequency channels.
\vspace{-1mm}
\end{enumerate}




The rest of the paper is  organized as follows. Section~\ref{sec:kernel} introduces our proposed kernel-based speech enhancement system: kernel machines, exponential power kernel, automatic hyper-parameter selection for subband adaptive kernels.  Experimental results and time complexity comparisons are discussed in Section~\ref{sec:results}. Section~\ref{sec:conclusion} gives the conclusion.


\section{KERNEL-BASED SPEECH ENHANCEMENT}
\label{sec:kernel}
\subsection{Kernel Machines}
The standard kernel methods for classification/regression denote a function $f$ that minimizes the discrepancy between $f(\bm{x}_j)$ and $y_j$, given labeled samples ${(\bm{x}_j, y_j)}_{j=1,...,n}$ where $\bm{x}_j \in \RR^d$ is a feature vector and $y_j\in \RR$ is its label.

Specifically, the space of $f$ is a Reproducing Kernel Hilbert Space $\HH$ associated to a positive-definite kernel function $k: \RR^d \times \RR^d \rightarrow \RR$.
We typically seek a function $f^* \in \HH$ for the following optimization problem:
\begin{equation}\label{eq:opt}
f^\star = \textnormal{argmin}_{f(\bx_j) = y_j,j=1,2,...,n}\norm{f}_\HH, 
\end{equation}
According to the Representer Theorem \cite{scholkopf2002learning}, $f^*$ has the form 
\begin{equation}
f(\bm{x}) = \sum_{j=1}^{n} \alpha_j k(\bm{x}, \bm{x}_j),
\end{equation}

To compute $f^*$ is equivalent to solve the linear system,
\begin{equation}
K\bm{\alpha}= (y_1, \cdots, y_n)^T,
\end{equation}
where the kernel matrix $K$ has entry
$[K]_{ij} = k(\bx_i, \bx_j)$
and $\balpha \defeq (\alpha_1, \cdots, \alpha_n)^T$ is the representation of $f$ under basis $\{k(\cdot, \bx_1), \cdots, k(\cdot, \bx_n) \}$.

\subsection{Exponential Power Kernel}

We use an exponential power kernel of the form
\begin{equation}
k_{\gamma, \sigma}(\bm{x}, \bm{z}) = \textnormal{exp}(-\frac{\lVert \bm{x}-\bm{z}\rVert ^ \gamma}{\sigma})
\end{equation}
for our kernel machine, where $\sigma$ is the kernel bandwidth and $\gamma$ is often called shape parameter.
\cite{giraud2006positive}
shows that the exponential power kernel is positive definite, hence a valid reproducing kernel.
This kernel also covers a large family of reproducing kernels including Gaussian kernel ($\gamma = 2$)
and Laplacian
kernel ($\gamma = 1$).
We observe that in many noise settings of
speech enhancement, the best performance is
achieved using this kernel with shape parameter $\gamma \leq 1$, which is 
\begin{wrapfigure}{r}{.25\textwidth}
  \includegraphics[width=.25\textwidth]
  {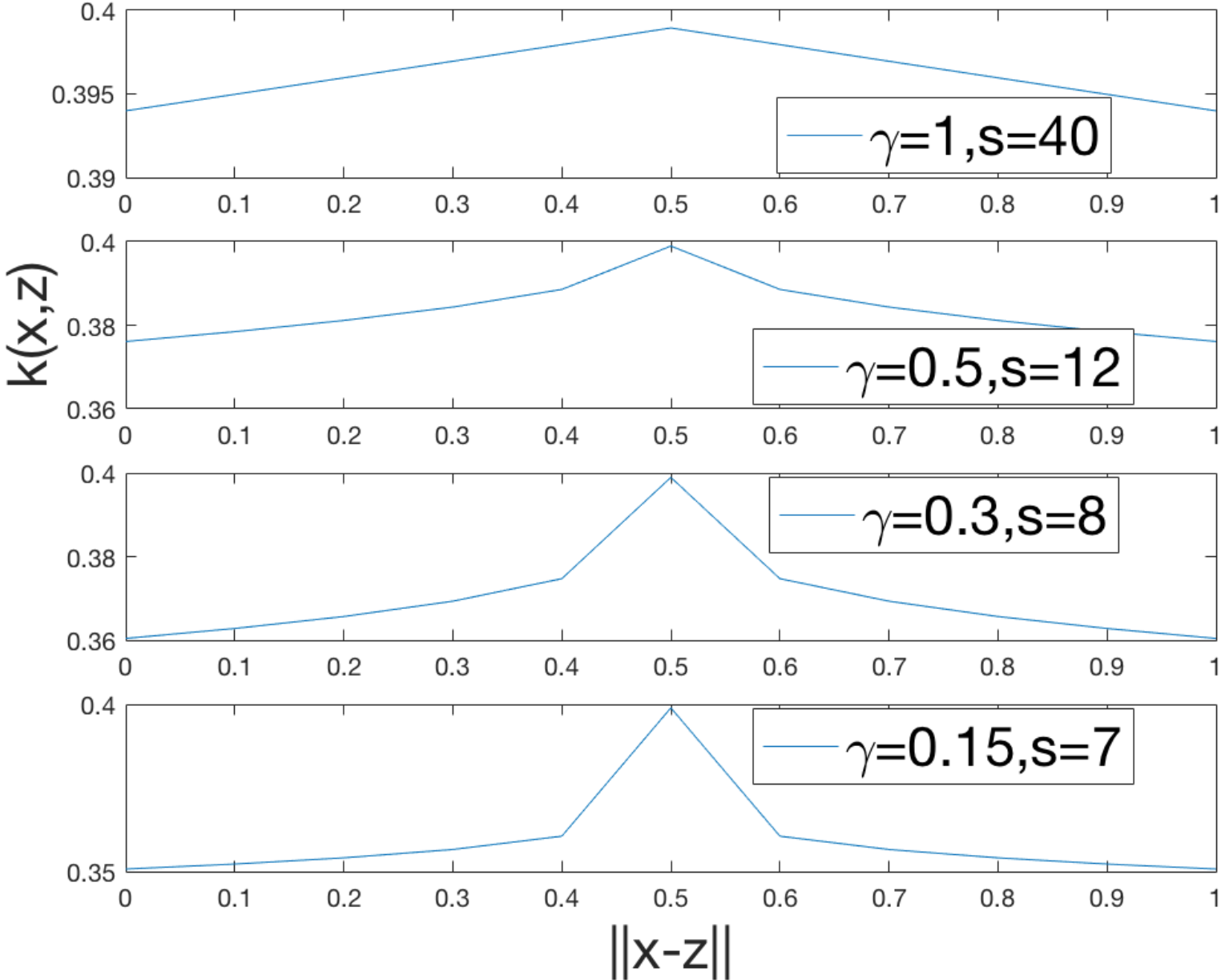}
\vspace{-8mm}
\end{wrapfigure}
highly non-smooth.
In the right side 
figure, we plot 
this kernel function with parameters that we use in our experiments.
We have not seen any application of this kernel (with $\gamma < 1$) in supervised learning literature.

\subsection{Automatic Subbands Adaptive Kernels}

\vspace{-3mm}
\begin{algorithm}
  \caption[c]{Automatic hyper-parameter selection\footnotemark}
  \label{alg:autotune}
  \begin{algorithmic}
     
    \State \textbf{Input}: $\bm{D}_{\tn{train}}, \bm{D}_{\tn{val}} $: training and validation data,
    $\Gamma$: a set of $\gamma$ for the exponential power kernel,
    $\sigma_{l}, \sigma_{h}$: smallest and largest bandwidth
    \State \textbf{Output}: selected kernel parameters $\gamma_{\tn{opt}}, s_{\tn{opt}}$ for $D_{\textnormal{train}}$

    \Procedure{\textnormal{autotune}}{$\bm{D}_{\tn{train}},\bm{D}_{\tn{val}}, \Gamma, \sigma_l, \sigma_h$}
    \State define {\bf subprocedure} cross-validate$(\gamma, \sigma)$ as: 
    train one kernel model with $k_{\gamma, \sigma}$ on $\bm{D}_{\tn{train}}$ using EigenPro iteration, return its loss on $D_{\tn{val}}$.

    \For  {$\gamma$ in $\Gamma$}
    \State $\sigma_{\gamma} = \textnormal{search}(\textnormal{cross-validate}(\gamma,\cdot), \sigma_l, \sigma_h)$
    \EndFor
    \vspace{-1.5mm}
    \State $\gamma_{\tn{opt}}, \sigma_{\tn{opt}} \leftarrow \textnormal{argmin}_{\gamma \in \Gamma, \sigma_\gamma}\textnormal{cross-validate}(\gamma,\sigma_\gamma)$

    \Return $\gamma_{\tn{opt}}, \sigma_{\tn{opt}}$
    \EndProcedure
    
    \Procedure{\textnormal{search}}{$f, \sigma_l, \sigma_h$}
    \If {($\sigma_h - \sigma_l \leq 2$)} 
    \State \Return $\sigma_l$
    \EndIf
    \vspace{-1.5mm}
    \State select $\sigma_{m1}, \sigma_{m2} \in (\sigma_l, \sigma_h)$
    \State compute $f(\sigma_l),
    f(\sigma_{m1}), f(\sigma_{m2}), f(\sigma_h)$
    
    \Switch{$\textnormal{min}\{f(\sigma_l), f(\sigma_{m1}), f(\sigma_{m2}), f(\sigma_h)\}$}
    \Case{$f(\sigma_l)$}:
    	\Return $\textnormal{search}(f, \sigma_l, \sigma_{m1})$
    \EndCase
    \Case{$f(\sigma_{m1})$}:
    	\Return $\textnormal{search}(f, \sigma_l, \sigma_{m2})$
    \EndCase
    \Case{$f(\sigma_{m2})$}:
    	\Return $\textnormal{search}(f, \sigma_{m1}, \sigma_h)$
    \EndCase
    \Case{$f(\sigma_h)$}:
    	\Return $\textnormal{search}(f, \sigma_{m2}, \sigma_h)$
    \EndCase
    \EndSwitch
    \EndProcedure
  \end{algorithmic}
\end{algorithm}
\vspace{-1mm}

\footnotetext{We apply memoization technique for computing cross-validate$(\cdot, \cdot)$.
We first attempt to set $\sigma_{m1}$, $\sigma_{m2}$ as a value that is already used in $(\sigma_l, \sigma_h)$, then we choose them to split $(\sigma_l, \sigma_h)$ into three parts as equal as possible.}

\begin{table*}[!ht]
\centering
\caption{Kernel \& DNN on TIMIT: (MSE: lowest is best, STOI and PESQ: highest is best. Best results bolded.)}
\label{tbl:timit}
\vspace{-1mm}
\resizebox{.9\textwidth}{!}{%
\begin{tabular}{|l|l||c|c|c||c|c|c||c|c|c|}
\hline
Noise &  \multirow{2}{*}{Metrics}    & \multicolumn{3}{c||}{5 dB}    & \multicolumn{3}{c||}{0 dB}    & \multicolumn{3}{c|}{-5 dB}   \\ \cline{3-11} 
 Type &   & Kernel & DNN & Noisy        & Kernel  & DNN  & Noisy       & Kernel    & DNN & Noisy        \\ \hline
\multirow{3}{*}{Engine}   & MSE ($\cdot 10^{-2}$) & \textbf{1.10}     & 1.41 & -      &\textbf{1.34} & 1.86 & -     & \textbf{1.17}  & 1.82  & - \\ \cline{2-11} 
                          & STOI                  & \textbf{0.91}    & 0.90 & 0.80   & \textbf{0.86} & 0.85 & 0.68  & \textbf{0.80}  & 0.77  & 0.57\\ \cline{2-11} 
                          & PESQ                  & \textbf{2.77} & \textbf{2.77} & 1.97    & \textbf{2.51}   & 2.45 & 1.66  & \textbf{2.19}  & 2.16 & 1.41  \\ \hline\hline
\multirow{3}{*}{Babble}   & MSE ($\cdot 10^{-2}$)   & \textbf{3.34}  & 3.49  & -      & \textbf{4.18}  & 4.37 & -    & \textbf{4.94}  & 5.43 & -\\ \cline{2-11} 
                          & STOI                 & \textbf{0.86}  & \textbf{0.86} & 0.77    & \textbf{0.77} & \textbf{0.77} & 0.66  & \textbf{0.64} & \textbf{0.64} & 0.55\\ \cline{2-11} 
                          & PESQ                   & \textbf{2.54} & 2.52 & 2.08 & \textbf{2.12} & 2.10  & 1.73  &\textbf{1.70} & 1.61  & 1.42  \\ \hline\hline
\multirow{3}{*}{SSN}      & MSE ($\cdot 10^{-2}$) & \textbf{1.35}     & 1.53 & -    & \textbf{1.48} & 1.67  & -    & \textbf{1.60}  & 1.76  & -\\ \cline{2-11} 
                          & STOI                   & \textbf{0.88} & \textbf{0.88} & 0.81    & \textbf{0.82} & \textbf{0.82} & 0.69  & \textbf{0.74} & \textbf{0.74} & 0.57\\ \cline{2-11} 
                          & PESQ                & \textbf{2.68}    & 2.66  & 2.05  & \textbf{2.36} & 2.32  & 1.75  & \textbf{2.03} & 2.00  & 1.48\\ \hline\hline
\multirow{3}{*}{Oproom}   & MSE ($\cdot 10^{-2}$)    & \textbf{1.44} & 1.85  & - & \textbf{1.34}     & 1.86  & -   &\textbf{1.17}  & 1.82  & - \\ \cline{2-11} 
                          & STOI                    & \textbf{0.88} & \textbf{0.88} & 0.79 & \textbf{0.84}  & 0.83  & 0.70  & \textbf{0.79} & 0.76  & 0.59 \\ \cline{2-11} 
                          & PESQ                 & \textbf{2.80}    & 2.79 & 2.16  & \textbf{2.50} & 2.47  & 1.78 & \textbf{2.23} & 2.12  & 1.40  \\ \hline\hline
\multirow{3}{*}{Factory1} & MSE ($\cdot 10^{-2}$)  & \textbf{2.51}   & 2.53  & -    & \textbf{2.52} & 2.55  & -    & \textbf{2.71}  & 2.77 & -  \\ \cline{2-11} 
                          & STOI                   & \textbf{0.86} & \textbf{0.86} & 0.77     & 0.78 & \textbf{0.79} & 0.65 & \textbf{0.68}  & \textbf{0.68} & 0.54  \\ \cline{2-11} 
                          & PESQ                  & \textbf{2.56}  & 2.51 & 1.99 & 2.20  & \textbf{2.23} & 1.62  & \textbf{1.79} & 1.77  & 1.29 \\ \hline
\end{tabular}
}
\vspace{-2mm}
\end{table*}

As empirically shown in Section~\ref{sec:single-multi}, we see that models with lower MSE at every frequency channel consistently outperform other models in STOI.
This motivates us to achieve lower MSE for all frequency channels by tuning kernel parameters for each of them.
In practice, we split the band of frequency channels into several blocks
, which we call {\it subband}s.

We propose a simple kernel-based framework as depicted in Fig.~\ref{fig:framework} to achieve automatic parameter selection and fast training for each subband. For $i$-th subband, the framework learns one model $f^{(i)}$ related to an exponential power kernel $k^{(i)}$ with parameters automatically tuned for this subband,
\begin{equation}
f^{(i)}(\bm{x}) = \sum_{j=1}^{n} \alpha_j^{(i)} k^{(i)}(\bm{x}, \bm{x}_j).
\end{equation}
Our framework starts by splitting the training targets $Y_\tn{train}$ into subband targets $Y^{(1)}, \cdots, Y^{(b)}$.
For training data related to the
$i$-th subband $(X_\tn{train}, Y^{(i)})$, we perform fast and automatic kernel parameter selection using {\it autotune} (Algorithm~\ref{alg:autotune})
on its subsamples, which selects one exponential power kernel $k^{(i)}$ for this subband.
We then train a kernel model on $(X_\tn{train}, Y^{(i)})$ with kernel $k^{(i)}$ using EigenPro iteration proposed in~\cite{ma2018learning}.
It learns an approximate solution $\alpha^{(i)}$ (or $f^{(i)}$) for the optimization problem~(\ref{eq:opt}).
Our final kernel machine is then formed by $\{ f^{(1)}, \cdots, f^{(b)} \}$.





For any unseen data $\bx$, our kernel machine first computes estimated mask $f^{(i)}(\bx)$ for each subband.
Then it combines the results of $\{f^{(1)}(\bx), \cdots, f^{(b)}(\bx)\}$ to obtain the estimated mask for all frequency channels.
Applying this mask to the noisy speech produces the estimated clean speech.

\section{EXPERIMENTAL RESULTS}
\label{sec:results}

We use kernel machines with 4 subbands (block of frequencies) for speech enhancement.
For fair comparison, we train both kernel machines and DNNs from scratch using the same features and targets.
We halt the training for any model when error on validation set stops decreasing.
Experiments are run on a server with 128GB main memory, two Intel Xeon(R) E5-2620 CPUs, and one GTX Titan Xp (Pascal) GPU.

\subsection{Regression Task}
\label{sec:regression}
We compare kernel machines and DNNs on a speech enhancement task described in~\cite{wang2014training} which is based on TIMIT corpus~\cite{garofolo1993darpa} and uses real-valued masks (IRM).
We follow the description in~\cite{wang2014training} for data preprocessing and DNN construction/training.
We consider five background noises: SSN, babble, a factory noise (factory1), a destroyer engine room (engine), and an operation room noise (oproom).
Every noise is mixed to speech at $-5, 0, 5$dB Signal-Noise-Ratio (SNR).


Table~\ref{tbl:timit} reports the MSE, STOI, and PESQ on test set for kernel machines and DNNs. We also present the STOI and PESQ of the noisy speech without enhancement.
For all noise settings, we see that kernel machines consistently produce better MSE, in many cases significantly lower than that from DNNs, which is also the training objective for both models.
We also see that STOI and PESQ of kernel machines are consistently better than or comparable to that from DNNs with only one exception (Factory1 0dB).

\subsection{Classification Task}
We train kernel machines and DNNs for a speech enhancement task in~\cite{healy2013algorithm}
which is based on HINT dataset and adopts binary masks (IBM) as targets.
We follow the same procedure described in~\cite{healy2013algorithm}
to preprocess the data and construct/train DNNs.
Specifically, we use two background noises, SSN and multi-talker babble. SSN is mixed to speech at -2, -5, -8dB SNR, and babble is mixed to speech at $0, -2, -5$dB SNR.
As our kernel machine is designed for regression task, we use a threshold $0.5$ to map its real-value prediction to binary target $\{0, 1\}$.

\begin{table}[htb]

\centering
\caption{Kernel \& DNN on HINT}
\label{tbl:hint}
\vspace{-2mm}
\resizebox{\columnwidth}{!}{%

\begin{tabular}{|l|l||l|l|l||l|l|l|}
\hline
\multirow{2}{*}{Metrics} & \multirow{2}{*}{Model} & \multicolumn{3}{c|}{Babble}                   & \multicolumn{3}{c|}{SSN}                      \\ \cline{3-8} 
                         &                        & 0dB           & -2dB          & -5dB          & -2dB          & -5dB          & -8dB          \\ \hline
\multirow{2}{*}{Acc}     & DNN                    & 0.90          & 0.91          & 0.90          & 0.91          & \textbf{0.91} & \textbf{0.92} \\ \cline{2-8} 
                         & Kernel                 & \textbf{0.92} & \textbf{0.92} & \textbf{0.91} & \textbf{0.92} & 0.90          & 0.89          \\ \hline
                         \hline
\multirow{2}{*}{STOI}    & DNN                    & 0.83          & 0.80          & 0.76          & 0.79          & \textbf{0.76} & \textbf{0.74} \\ \cline{2-8} 
                         & Kernel                 & \textbf{0.86} & \textbf{0.83} & \textbf{0.78} & \textbf{0.81} & 0.75          & 0.71          \\ \hline
\end{tabular}

}
\vspace{-2mm}
\end{table}

In Table~\ref{tbl:hint}, we compare the classification accuracy (Acc) and STOI of kernel machine and DNNs under different noise settings. We see that our kernel machines outperform DNNs on noise settings with babble and perform worse than DNN on noise settings with SSN.
In all, the proposed kernel machines match the performance of DNNs on this classification task.

\subsection{Single Kernel and Subband Adaptive Kernels}
\label{sec:single-multi}
We start by analyzing the performance of 
kernel machines that use a single kernel for all frequency channels on the regression task in Section~\ref{sec:regression}. The training of such kernel machine (1 subband) is significantly faster than that of our default kernel machine (4 subbands).
Remarkably, its performance is also quite competitive. It consistently outperforms DNNs in MSE in all noise settings. In 8 out of 15 noise settings, it produces STOI the same as that from the kernel machine with 4 subbands (it also produces nearly same PESQ).
\begin{table}[!ht]
\centering
\vspace{-3mm}
\caption{Comparison of kernel machines with 1 subband and 4 subbands}
\label{tbl:single-multi}
\vspace{-2mm}
\begin{adjustbox}{center}
\resizebox{\columnwidth}{!}{
\begin{tabular}{|l|l|c|c|c|}
\hline
\multicolumn{1}{|l|}{\begin{tabular}[c]{@{}c@{}} Noise \\ setting \end{tabular}}  & Metrics                                                    & \begin{tabular}[c]{@{}c@{}}Kernel \\(1 subband) \end{tabular} & \begin{tabular}[c]{@{}c@{}} Kernel \\(4 subbands) \end{tabular} & DNN           \\ \hline
\multirow{3}{*}{\begin{tabular}[c]{@{}l@{}}SSN\\ 0dB\end{tabular}}       & MSE  & 1.60                                                            & \textbf{1.48}                                                   & 1.67          \\ \cline{2-5} 
                                                                         & STOI & 0.81                                                            & \textbf{0.82}                                                   & \textbf{0.82} \\ \cline{2-5} 
                                                                         & PESQ & 2.35                                                            & \textbf{2.36}                                                   & 2.32          \\ \hline
                                                                         \hline
\multirow{3}{*}{\begin{tabular}[c]{@{}l@{}}SSN\\ -5dB\end{tabular}}      & MSE  & 1.67                                                            & \textbf{1.60}                                                   & 1.76          \\ \cline{2-5} 
                                                                         & STOI & 0.73                                                            & \textbf{0.74}                                                   & \textbf{0.74} \\ \cline{2-5} 
                                                                         & PESQ & 2.01                                                            & \textbf{2.03}                                                   & 2.00          \\ \hline
                                                                         \hline
\multirow{3}{*}{\begin{tabular}[c]{@{}l@{}}Factory1\\ -5dB\end{tabular}} & MSE  & 2.76                                                            & \textbf{2.71}                                                   & 2.77          \\ \cline{2-5} 
                                                                         & STOI & 0.67                                                            & \textbf{0.68}                                                   & \textbf{0.68} \\ \cline{2-5} 
                                                                         & PESQ & 1.78                                                   & \textbf{1.79}                                                            & 1.77 \\ \hline
\end{tabular}
}
\end{adjustbox}
\vspace{-2mm}
\end{table}

However, in other noise settings, kernel (1 subband) has smaller training loss (MSE) than DNNs, but no better STOI (we show three cases in Table~\ref{tbl:single-multi})~\cite{lightburn2016weighted,zhang2018training}. To improve desired metrics (STOI/PESQ), we first compare the MSE of every frequency channel of DNNs and kernel machines. 
\begin{figure}[htb]
\centering
\begin{subfigure}{.25\textwidth}
  \centering
  \includegraphics[width=1\linewidth]{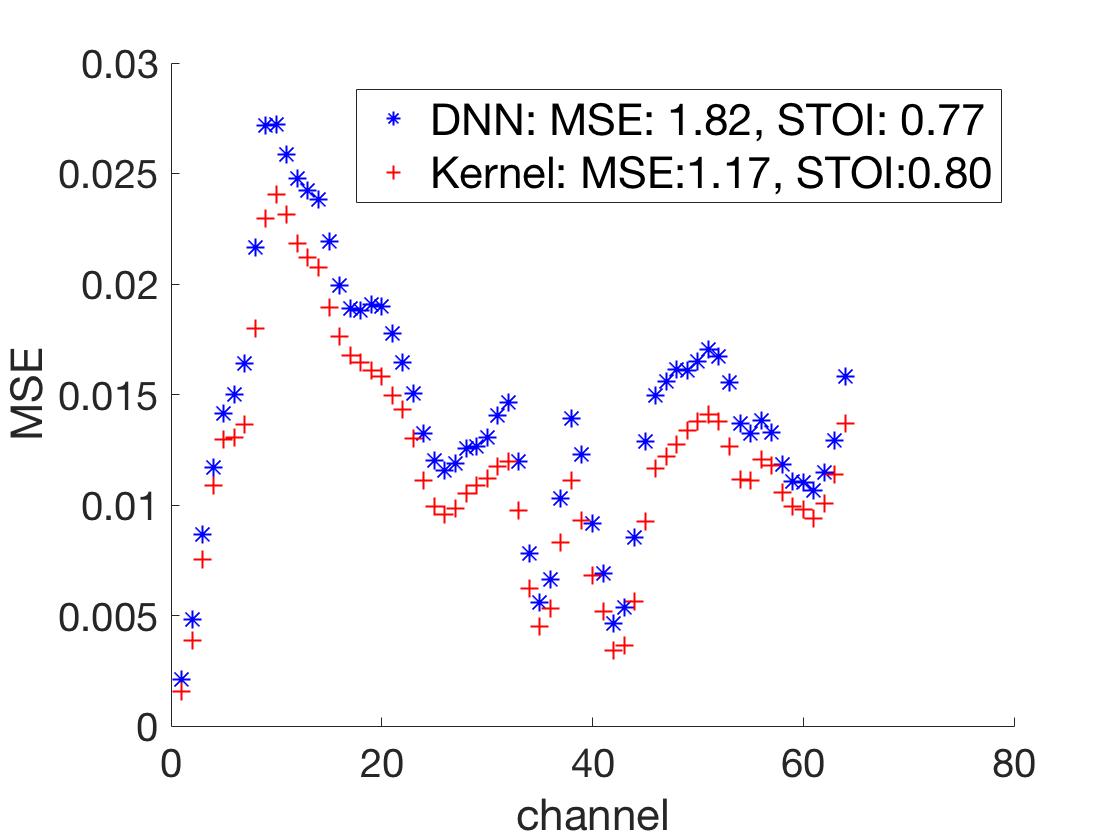}
  \vspace{-5mm}
  \caption{Engine -5dB}
  \label{fig:engine-5}
\end{subfigure}%
\begin{subfigure}{.25\textwidth}
  \centering
  \includegraphics[width=1\linewidth]{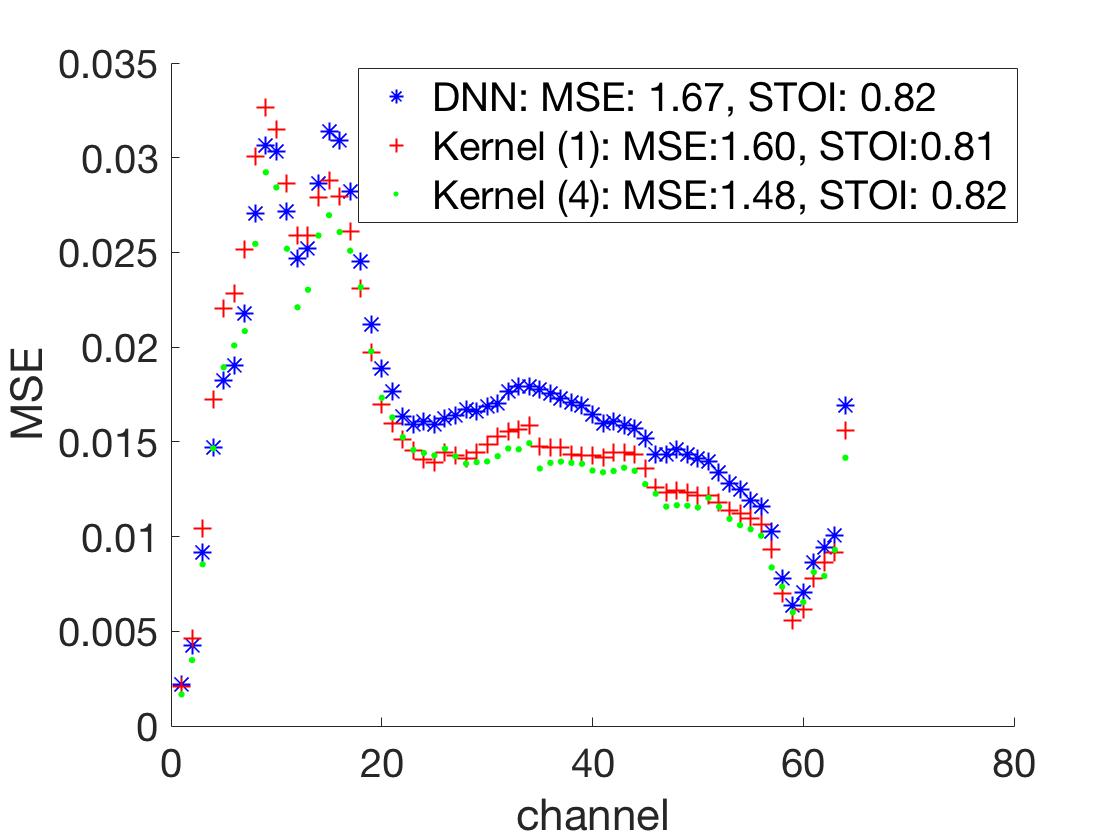}
  \caption{SSN 0dB}
  \label{fig:ssn0}
\end{subfigure}
\vspace{-3mm}
\caption{MSE along per frequency channel}
\vspace{-3mm}
\end{figure}

As shown in Fig.~\ref{fig:engine-5},  for cases that kernels have much smaller overall MSE and smaller MSE on each frequency channel, kernels also achieve better STOI. For cases like SSN 0dB, as shown in Fig.~\ref{fig:ssn0}, even though single kernel (1 subband) has smaller overall MSE, its STOI is not as good as DNNs. Multiple kernels (4 subbands) decrease MSE further and also achieve better STOI. This shows that having smaller MSE along all frequency channels leads to better STOI. This reveals a correlation between MSE and STOI/PESQ associated with frequency channels.

\subsection{Time Complexity}
\begin{table}[!ht]
\centering
\vspace{-3mm}
\caption{Running time/epochs of Kernel \& DNN}
\label{tbl:complexity}
\vspace{-2mm}
\begin{adjustbox}{center}
\resizebox{\columnwidth}{!}{
\begin{tabular}{|l||c|c|c||c|c|}
\hline
\multirow{3}{*}{Dataset} & \multicolumn{3}{c||}{Time (minutes)}                & \multicolumn{2}{c|}{Epochs}                    \\ \cline{2-6} 
                        & \multicolumn{2}{c|}{Kernel} & \multirow{2}{*}{DNN} & \multirow{2}{*}{Kernel} & \multirow{2}{*}{DNN} \\ \cline{2-3}
                        & 1 subband       & 4 subbands      &                      &                         &                      \\ \hline
HINT                    & 0.8          & 3.2          & 6.6                  & 10                      & 50                   \\ \hline
\hline
TIMIT                   &      18        &     65         &   124                   & 5                       & 93                   \\ \hline
\end{tabular}

}

\end{adjustbox}
\vspace{-2mm}
\end{table}

In Table~\ref{tbl:complexity}, we compare the training time of DNNs and kernel machine on both HINT and TIMIT.
Note that the training of kernel machines in all experiments typically completes in no more than 10 epochs, significantly less than the number of epochs required for DNNs.
Furthermore, the training time of kernel machines is also less than that of DNNs. Notably, training kernel machine with 1 subband takes much less time than DNNs.


\section{CONCLUSION}
\label{sec:conclusion}

In this paper, 
we have shown that kernel machines using exponential power kernels show strong performance on speech enhancement problems.
Notably, our method needs no parameter tuning for optimization and employs nearly automatic tuning for kernel hyper-parameter selection.
Moreover, we show that the training time and computational requirement of our method are comparable or less than  those needed to train  neural networks.
We expect that this highly efficient kernel method will be useful for other problems in speech and signal processing.

\vfill\pagebreak

\label{sec:refs}

\bibliographystyle{IEEEbib}
\bibliography{strings,refs}

\end{document}